\documentclass[letterpaper, 10 pt, conference]{ieeeconf}  

\IEEEoverridecommandlockouts
\usepackage{svg}
\usepackage{algorithm}
\usepackage{algorithmicx}
\usepackage{algpseudocode}
\usepackage{amsmath}
\usepackage{subcaption}
\usepackage{placeins}
\usepackage{cite}

\title{\LARGE \bf
Robotic Exploration through Semantic Topometric Mapping
}

\author{Scott Fredriksson, Akshit Saradagi and George Nikolakopoulos
\thanks{Scott Fredriksson is the corresponding author (email : scofre@ltu.se)
The authors are with the Robotics and AI group, in the Department of Computer Science, Electrical and Space Engineering at Luleå University of Technology, Sweden.}%
}

\begin{document}

\maketitle
\thispagestyle{empty}
\pagestyle{empty}

\begin{abstract}
In this article, we introduce a novel strategy for robotic exploration in unknown environments using a semantic topometric map. As it will be presented, the semantic topometric map is generated by segmenting the grid map of the currently explored parts of the environment into regions, such as intersections, pathways, dead-ends, and unexplored frontiers, which constitute the structural semantics of an environment. The proposed exploration strategy leverages metric information of the frontier, such as distance and angle to the frontier, similar to existing frameworks, with the key difference being the additional utilization of structural semantic information, such as properties of the intersections leading to frontiers. The algorithm for generating semantic topometric mapping utilized by the proposed method is lightweight, resulting in the method's online execution being both rapid and computationally efficient. Moreover, the proposed framework can be applied to both structured and unstructured indoor and outdoor environments, which enhances the versatility of the proposed exploration algorithm. We validate our exploration strategy and demonstrate the utility of structural semantics in exploration in two complex indoor environments by utilizing a Turtlebot3 as the robotic agent. Compared to traditional frontier-based methods, our findings indicate that the proposed approach leads to faster exploration and requires less computation time.
\end{abstract}
%
%
\begin{figure*}[t]
    \centering
    \vspace*{6px}
    \includegraphics[width=.9\linewidth]{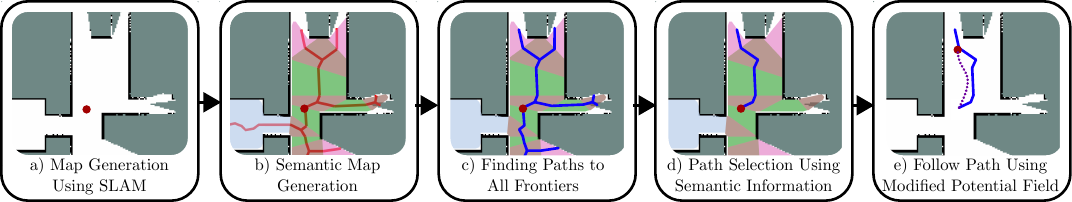}
    \caption{An illustration of the steps taken by the proposed method to explore an unknown environment. The proposed method processes the occupancy map of the currently explored environment (Fig. 1a)), to find the optimal goal location for further exploration (Fig. 1d)), the path to which is smoothly tracked by the robot using modified potential fields (Fig. 1e)).}
    \label{fig:method}
\end{figure*}
\section{INTRODUCTION}
Robot exploration, often referred to as Active Simultaneous Localization and Mapping (ASLAM), is an automated version of the Simultaneous Localization and Mapping (SLAM) process \cite{Placed2023}. ASLAM holds significant importance in mobile robotics due to several reasons. Firstly, it enables a robot to autonomously construct its own map of the operating area. Secondly, it becomes particularly essential in situations where manual control of the robot is impractical, such as in subterranean environments or during search and rescue missions.

The autonomous exploration problem is, in general, divided into three distinct components: pose identification, optimal goal selection, and navigation~\cite{Lluvia2021}. Pose detection involves identifying potential locations on the map that the robot can visit to acquire new information about its environment. Optimal goal selection, as the name suggests, is the process of choosing the pose that maximizes both current and future information gain. Lastly, the navigation component safely directs the robot to the chosen goal position.

The majority of the methods in ASLAM focus on the concept of frontiers, which are collections of unoccupied cells on a map adjacent to unexplored regions. Conventional techniques involve identifying these frontiers by locating their boundaries on the map~\cite{Yamauchi1997, Topiwala2018, Sun2020}. However, this often necessitates scanning large parts of the map, which can be computationally intensive, especially as the map expands in size. 
An alternative common approach, which is less computationally demanding, is to utilize Rapidly-exploring Random Trees (RRTs) to detect frontiers~\cite{Umari2017, Tran2023}. A closely related approach to RRTs involves the utilization of a Voronoi graph~\cite{Aurenhammer1991} to direct the exploration, as demonstrated in~\cite{Li2020}.

Frontier-based exploration has limitations due to its simplistic view of the world as comprising just the regions in between explored and unexplored areas of the map. Therefore, in this article, instead of attempting to address the exploration problem from a frontier-centric perspective, we look to explore the problem from a more semantically meaningful perspective. By representing the world as a collection of intersections, dead ends, and pathways, some of which will lead to more unexplored regions on the map, we leverage this structural semantic information for exploration.
\subsection{Related work}
Using semantic information in exploration is a relatively new idea that is likely going to become more popular in the future due to advancements in the fields of semantic mapping~\cite{Kostavelis2015} and scene graphs~\cite{Chang2023}. 
A notable instance of this approach is highlighted in the research of~\cite{Cowley2011}, where the explored map was converted to a topometric map divided into semantic "places" that subsequently aid in determining optimal goal positions. However, this approach poses limitations since it presumes that all walls in the map are oriented at \(90^{\circ}\) angles relative to each other, which limits the types of environments in which the framework can operate. 
Another example of the use of semantic information utilized in exploration can be found in~\cite{Asgharivaskasi2023}, where semantic labels are integrated into an OcTree map to enhance the exploration process. Their approach uses detailed semantic labels, such as "tree," "building," "grass," and so forth.  Similarly, the work in \cite{Chen2023} integrates semantic objects into frontiers to locate objects in home environments quickly. In contrast, our proposed framework utilizes fundamental semantic information, emphasizing elements like pathways and intersections.

Another approach, similar to utilizing semantic information for exploration, is the integration of machine learning to enhance robotic exploration. The study presented in~\cite{Shrestha2019} employs a generative neural network to predict unmapped areas, thereby optimizing goal selection during the exploration phase.
The work in~\cite{Tao2023} used self-supervised machine learning to do information gain prediction on a 3D occupancy grid and detection of semantic objects to improve the exploration process for micro aerial vehicles.

In contrast to the related literature, in this article, we introduce a novel approach to autonomous robot exploration that utilizes structural semantic information through semantic topometric mapping. 
\subsection{Contributions}
Firstly, we introduce a fundamentally novel and distinct approach to autonomous robotic exploration, wherein structural semantic information contained in the currently explored map, such as properties of intersections, pathways, dead ends, and pathways leading to frontiers, are incorporated to enhance the exploration process. Such semantic information has not been harnessed so far in the literature relating to robotic exploration to the best of the author's knowledge. The main advantages of the proposed method, when compared to e.g.~\cite{Yamauchi1997, Topiwala2018, Sun2020,Umari2017, Tran2023}, stems from the pre-processing stage of the method, where the already explored map is segmented to derive structural semantics. This stage supplies the necessary information for all three stages of exploration: pose identification (unexplored frontiers), optimal goal selection (incorporation of properties of intersections leading to a frontier), and efficient navigation (paths derived from the global skeleton map). In the presented approach, the necessary structural and topological properties are derived from a map produced by LiDAR measurements, as against visual semantics in ~\cite{Asgharivaskasi2023} that require an additional sensor. 
Finally, the proposed framework is comprehensively compared to an existing frontier-based exploration framework \cite{Topiwala2018} through real-world experiments, which showed that the proposed method completes exploration faster than the framework in \cite{Topiwala2018} while requiring significantly lower computational times for map generation (Figure \ref{fig:MapGeneration}), map exploration (Figure \ref{fig:MapExplorationTime}) and global navigation (Figure \ref{fig:GlobalNavigation}). 
%
%
\section{Methodology}
In this Section, we present details of the autonomous robotic exploration strategy proposed in this article, with the overall concept illustrated in Figure \ref{fig:method}.
\subsection{Pose Identification}
\begin{figure}[b]
    \centering
    \includegraphics[width=\linewidth]{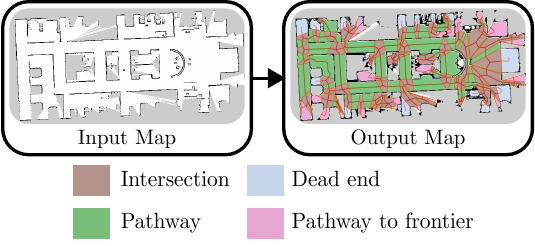}
    \caption{An example of a semantic topological map produced by the method in \cite{fredriksson2023semantic}.}
\label{fig:GFmethod}
\end{figure}

Similar to other exploration frameworks, the proposed method utilizes frontiers as potential exploration poses. The key difference that sets our proposed approach apart is that the frontiers are derived as part of a semantic topometric map produced by the framework introduced in~\cite{fredriksson2023semantic}. This framework processes the currently explored 2D grid-based map, $\mathcal{M}$, originating from a SLAM solution to yield a semantic topometric map $\mathcal{S}$. Figure \ref{fig:GFmethod} presents an illustrative example of the semantic segmentation performed by the framework proposed in ~\cite{fredriksson2023semantic}. A topometric map is a hybrid representation, bridging the gap between a grid-based metric map and a topological map \cite{Remolina2004}. The semantic map created by the framework comprises a set of semantic areas $\mathcal{A}$, which correspond to various semantic types such as intersection, pathway, dead-end, and paths that lead to unexplored areas. Additionally, the framework outputs a set of paths represented as a skeleton map of the environment, which can aid a global navigator in swiftly and efficiently navigating the map. Among the semantic areas are $N_f$ pathways leading to frontiers, collected into $\mathcal{A}_F\subset\mathcal{A}$. To each $A^F_{i}\in\mathcal{A}_F$, $i\in\{1, \ldots, N_f\}$, there is a global path $P_i$ that connects the robot's current pose to the associated unexplored frontier $F_i$. The set $\mathcal{F}=\{F_i\}$ is then the collection of all target poses for robot exploration. 
\subsection{Optimal Goal Selection} \label{sec:goalSel}
\begin{figure}[t!]
    \centering
    \vspace*{6px}
    \includegraphics[width=0.8\linewidth]{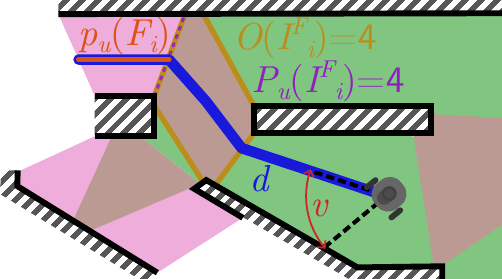}
    \caption{Illustration of a path leading to a frontier in a semantic topometric map, highlighted in blue, for a specific scenario. The figure also shows the different semantic entities used in the optimal selection process presented in Section \ref{sec:goalSel}. The intersection connected to the target frontier has four openings, and only one of its pathways leads directly to a frontier.}
    \label{fig:goalValues}
\end{figure}

A common approach for goal selection, adopted by many exploration frameworks, is the use of the metric information of the frontier, distance, and angles to determine the optimal goal position $F^*\in \mathcal{F}$. This is often done without taking the map into consideration, primarily because it is computationally expensive to calculate paths to every frontier on the map. In contrast, when utilizing the semantic topometric map, paths can be found to all possible poses relatively computationally cheaply, as described in subsection~\ref{sec:nav}. Therefore, the distance cost is determined by the length $d$ of the path from the robot to each pose, and the angle cost $v$ is determined by the difference between the robot's current pose and the pose at the start of the global path to the frontier, illustrated in Figure \ref{fig:goalValues}.

In addition to the length $d$ and angle cost $v$, semantic information from the topometric map is employed in the proposed method to determine the optimal exploration goal $F^*$. The first consideration is the length $p_l(F_i)$ of the path between a frontier $F_i$ and the nearest intersection denoted as $I^F_i$. This is motivated by the observation that paths leading to unexplored regions, which don't lead anywhere (such as walls hidden behind other walls at intersections), are typically very short.  
The proposed method also uses the properties of the intersection connected to the path leading to the unexplored area $I^F_i$. Specifically: 1) The number of openings, $O(I^F_i)$, in the intersection $I^F_i$, with the assumption that an intersection with numerous openings is significant in the overall map, and hence, exploring paths connected to it will yield higher information gain than those linked to intersections with fewer openings and 2) The number of pathways $P_u(I^F_i)$, connected to the intersection directly leading to a frontier, as a higher number of connected pathways suggests higher future information gain, resulting in lesser backtracking for a robot.

The equation below consolidates all the above cost and gain considerations and presents a single metric for each frontier $F_i$:
\begin{equation}
    \label{eq:pathCost}
        C(F_i) = C_{metric}-G_{semantic} 
\end{equation}
where
\begin{equation}
    \begin{aligned}
        C_{metric}&=\dfrac{d}{L_{path}} + w_v \cdot v \vspace{5px} \\
        G_{semantic}&=w_p \cdot p_{l}(F_i) + w_I \cdot (O(I^F_i) + P_u(I^F_i)).
    \end{aligned}
    \label{eq:cost_split}
\end{equation}
The terms in Equations \eqref{eq:pathCost} and \eqref{eq:cost_split} are defined as follows: \( C_{metric} \) is the cost of the path using the metric properties of the path. \( G_{semantic} \) measures the semantic value gain of the path based on the semantic properties along the path leading to the frontier. 
Finally, \( w_v \), \( w_p \), and \( w_I \) are weights set by the user to determine the importance of the various factors.
The weighting parameters are tuned based on the following rationale: 

\subsubsection{Tuning of $w_v$} A higher value of \( w_v \) decreases the likelihood that the robot will turn around, thereby committing the robot to the direction in which it is currently headed. This is advantageous for avoiding local cost minima, but an excessively high value of \( w_v \) may cause the robot to make poor decisions, leading it to take unnecessary detours to avoid turning around.

\subsubsection{Tuning of \( w_p \) and \( w_I \)}  Low values for \( w_p \) and \( w_I \) are preferred to ensure that the robot does not bypass nearby frontiers in favor of more distant but complex frontiers. However, \( w_p \) and \( w_I \) should be sufficiently high to allow the robot to choose a nearby frontier with a higher potential for information gain when faced with multiple nearby options.

The variable \( L_{path} \) represents the average length of all paths between intersections and serves as a normalizing factor. In more complex environments where the distances between intersections are small, the weights \( w_v \), \( w_p \), and \( w_I \) become less critical because the distance \( d \) is generally lower. Conversely, in larger maps with greater distances between intersections, both the distance \( d \) and the weights are likely to be higher. This ensures that the robot's behavior remains consistent regardless of the complexity of the environment.
With the cost associated with each frontier $F_i$ as defined in \eqref{eq:pathCost}, the optimal goal selection reduced to the following optimization problem. 
\begin{equation}
    \label{eq:Fstar}
        F^*=\arg \min_{F_i\in \mathcal{F}}(C(F_i)) 
\end{equation}
The algorithm to find the path $P$ that connects the robot's current pose to $F^*$, from the skeleton map of the environment, is presented in the next subsection.  
\subsection{Navigation} \label{sec:nav}
The proposed Algorithm \ref{alg:PathFinding} outlines the method used to find the optimal path, utilizing the cost function from Eq. \eqref{eq:pathCost}. The cost is applied at two points in the algorithm; the angle cost $w_v*v$ is calculated at the start of each path (at line 3), and the remaining costs are calculated after finding the complete path (at line 9). To optimize the search process, the algorithm incorporates one additional condition. The condition specifies that if a path reaches an area that another path has already reached with a lower score, the current path is discontinued (The current value of each area is tracked by the $A_{cost}$ array).

Utilizing pre-calculated global paths in the semantic topometric map is computationally efficient. However, as these paths are general global paths generated without any consideration of the robot's current position, they are typically suboptimal. This is true not only in terms of the shortest point-to-point maneuver but also with respect to smoothness and length (see Figure \ref{fig:method}b)). To mitigate these shortcomings, we propose a novel local navigation scheme using a modified version of a potential field.

We consider a robot equipped with a 2D LiDAR sensor that receives a set of \( N \) range measurements, denoted as \( \mathcal{L}=\{L_1, L_2,\ldots, L_i, \ldots, L_N\} \), during each update. Let \( \mathcal{L}_v = \{L_{v1}, L_{v2}, \ldots, L_{vi}, \ldots, L_{vN}\} \) represent the angles corresponding to each LiDAR range measurement. The potential field is described by the following equation:

\begin{equation}
    \label{eq:potenField}
    \Bar{F} = f_g
    \begin{bmatrix}
    \cos(G_v)\\
    \sin(G_v)
    \end{bmatrix}
    - \sum_{i=1}^{N} \frac{f_o}{L_i}
    \begin{bmatrix}
    \cos(L_{vi})\\
    \sin(L_{vi})
    \end{bmatrix}
\end{equation}

\begin{figure}[t]
    \centering
    \vspace*{6px}
    \includegraphics[width=.7\linewidth]{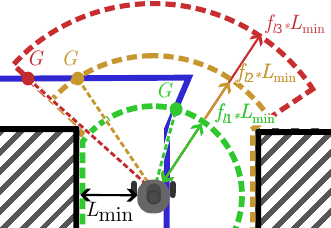}
    \caption{Illustration of the goal selection for point \( G \) on the path represented as a blue line. The area detectable by the LiDAR, up to a range of \( f_l*  L_{\text{min}} \) for different values of \(f_l\) (\(f_{l1}\), \(f_{l2}\) and \(f_{l3}\)), is marked by dotted lines.}
    \label{fig:lmin}
\end{figure}

In Eq. \eqref{eq:potenField}, \( \Bar{F} \) is the motion vector relative to the robot. The parameters \( f_g \) and \( f_o \) are tunable factors affecting the potential field. The angle \( G_v \) is the direction towards the goal position \( G \), which is a point on the global path that is not obscured by any obstacles and is within a distance of \( f_l*L_{\text{min}} \), where \( L_{\text{min}} = \min(\mathcal{L}) \) and \( f_l \) is a scaling factor, as illustrated in Figure \ref{fig:lmin} that shows the effect of different choices of \( f_l \).
\begin{figure}[t]
\begin{algorithm}[H]
\caption{Method To Find Optimal Path}
\label{alg:PathFinding}
\begin{flushleft}
\hspace*{\algorithmicindent} \textbf{Input:} $firstNode$ : The semantic area robot is in.\\
\hspace*{\algorithmicindent} \textbf{Initiate:} $searchList$ : The paths being searched.\\ 
\hspace*{53px} $candidate$ : the current best path. \\
\hspace*{53px} $A_{cost}$ : array of size $\mathcal{A},$ set to $\infty$
\end{flushleft}
\begin{algorithmic}[1]
\For{$N \in get\_neighbour (firstNode)$}
    \State Compute $v$ from robot to path to $N$ 
    \State Add new $N$ to $searchList$ with computed cost $w_v*v$
\EndFor

\While{$searchList \neq \emptyset$}
    \State $c_{path} \gets$ last element of  $searchList$
    \State Remove $c_{path}$ from $searchList$
    \If{$c_{path} \in \mathcal{F}$}
        \If{$cost(c_{path})<candidate.score$}
            \State Update $candidate$
        \EndIf
    \EndIf
    \For{$N \in get\_neighbour (c_{path}.last)$}
        \State Compute $d$ between $c_{path}.last$ and $N$
        \State If $A_{cost}[N]<c_{path}.cost+d$ continue
        \State $c_{path} \gets c_{path}+N$
        \State $c_{path}.cost \gets c_{path}.cost + d$
        \State $A_{cost}[N] \gets c_{path}.cost$
        \State Add $c_{path}$ to $searchList$
    \EndFor
\EndWhile
\end{algorithmic}
\hspace*{\algorithmicindent} \textbf{Output:} $candidate$ : as a path $P$
\end{algorithm}
\end{figure}
The potential field is tuned using the following method: The value $f_o$ is increased until the robot does not collide with walls. The value $f_g$ is incremented until the robot doesn't get stuck in local force minima. The value $f_l$ determines the strictness of the path; a lower value will result in the robot closely following the path but may lead to suboptimal routes to reach the goal. 
This is illustrated by the value $f_{l1}$ in Figure \ref{fig:lmin}. Conversely, a higher value of \(f_l\) allows the robot to pursue faster, more optimal paths but might bring it closer to walls, increasing the risk of collisions, as illustrated by the value $f_{l3}$ in Figure \ref{fig:lmin}.
\section{Validation and comparison}
\subsection{Experimental Setup}
\begin{figure}[b!]
    \centering
    \begin{subfigure}{0.7\linewidth}
        \centering
        \includegraphics[width=\linewidth]{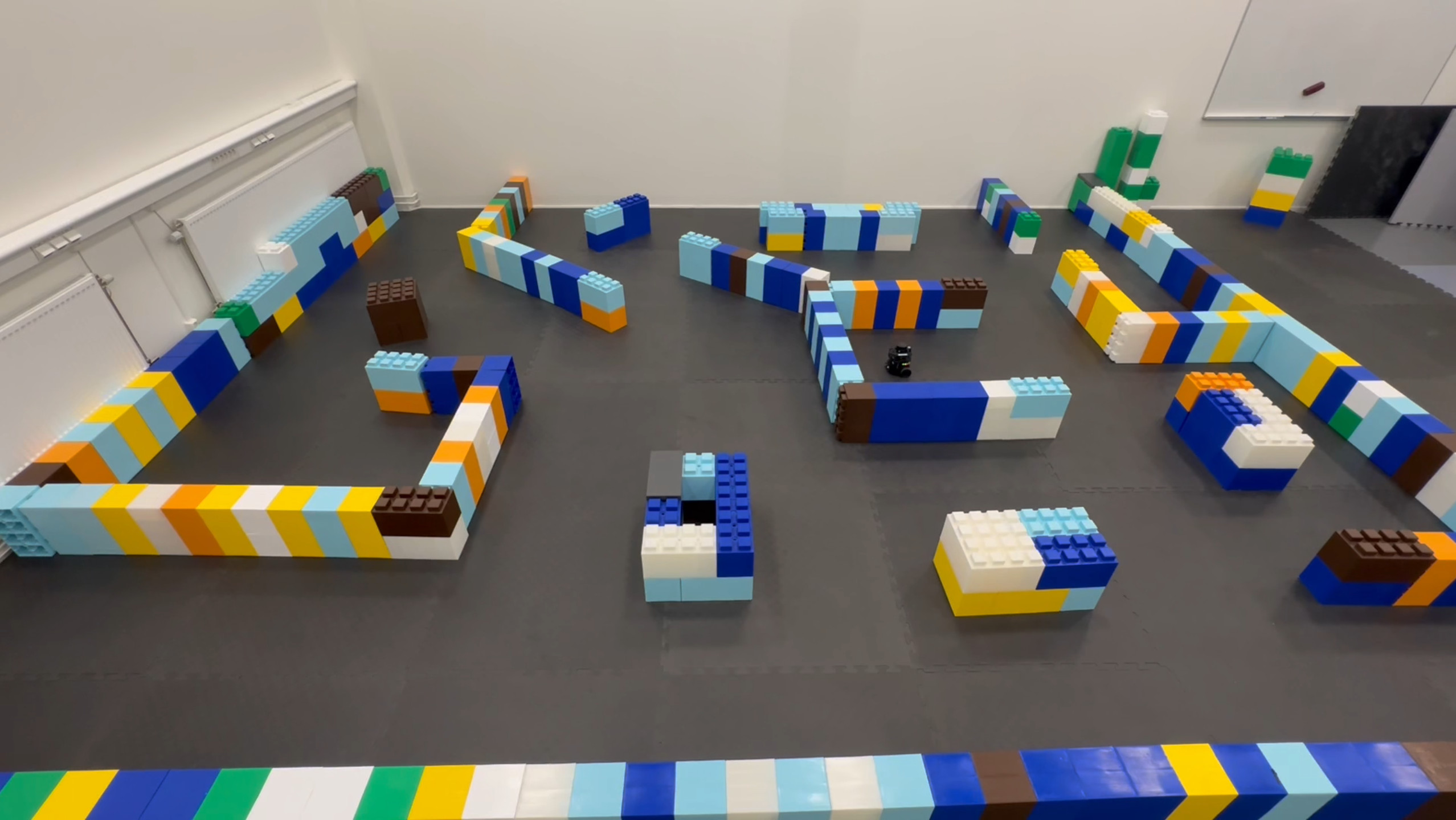}
        \caption{Maze Configuration 1}
        \label{fig:maze1}
    \end{subfigure}
    \begin{subfigure}{0.7\linewidth}
        \centering
        \includegraphics[width=\linewidth]{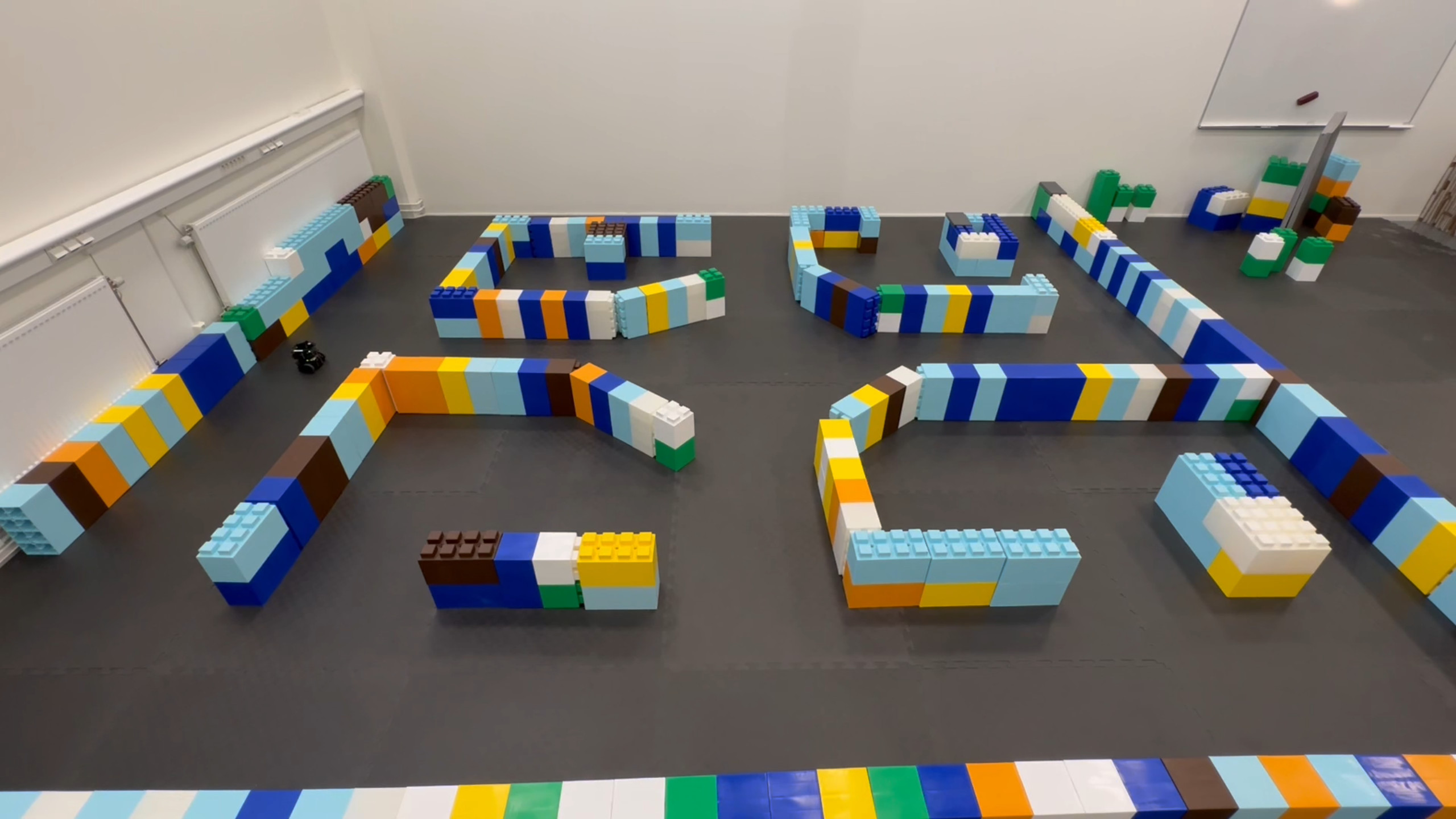}
        \caption{Maze Configuration 2}
        \label{fig:maze2}
    \end{subfigure}
    \caption{The two test environments used in the validation and testing.}
    \label{fig:maze}
\end{figure}
\begin{figure}
    \centering
    \vspace*{6px}
    \includegraphics[width=.9\linewidth]{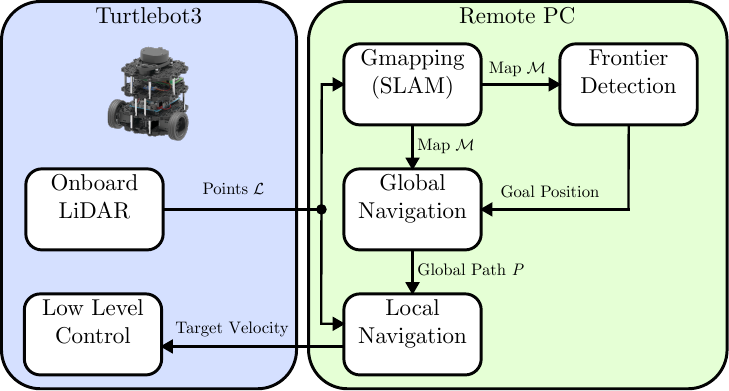}
    \caption{Robot Software Stack and implementation of the proposed exploration strategy.}
    \label{fig:RobotSetUp}
\end{figure}

The Proposed Method (PM) was benchmarked against an existing framework, \textit{frontier\_exploration} (FE) \cite{Topiwala2018}. The latter is a wavefront frontier
detector (WFD) \cite{keidar2012robot} based approach, which employs both frontier size and distance metrics for optimal goal selection. Both frameworks were evaluated in two distinct scenarios, utilizing the TurtleBot3 robot platform. The robot navigated through two maze configurations shown in figure \ref{fig:maze}, that were custom-built for the validation tests. For each maze configuration, 10 successful exploration missions were performed with both frameworks. Both frameworks were run until they self-stopped, which is when they deemed that the complete map had been explored (no more frontiers left to explore).
The software stack of the robot is detailed in Figure \ref{fig:RobotSetUp}. Both frameworks utilize \textit{Gmapping} \cite{Grisetti2007} as their SLAM method. The map resolution was chosen to be 0.05 m. FE uses \textit{Navfn} as its global planner and \textit{DWA} as its local planner, both of which are components of the navigation stack part of the Robot Operating System (ROS) \cite{quigley2009ros}. In contrast, PM uses the semantic topometric map \cite{fredriksson2023semantic} for frontier detection, with global navigation strategies and local navigation presented in subsection \ref{sec:nav}. Optimal goal selection, presented in subsection \ref{sec:goalSel}, is performed during the frontier selection process in the FE framework, whereas in PM, it is carried out in the global navigation step. Most of the ROS nodes were executed on a remote PC equipped with an AMD 5850u CPU and running Linux, kernel version 6.4.12.

To evaluate and compare the exploration performance of both methods, we use four metrics: a) the unoccupied area explored over time, b) the time required to complete the exploration process, c) the computational time for frontier detection, and d) the computational time for global navigation. In Tables \ref{tab:goal_select_parameters}, we list the parameters used in the PM. 
The Turtlebot3 robot platform's maximum heading speed was limited to 0.15 m/s, and maximum rotation speed was limited to 2 rad/s. 

\begin{table}[h]
    \centering
    \begin{tabular}{|c|c|c|c|c|c|c|}
    \hline
        Parameter & $w_v$ & $w_p$ & $w_I$ & $f_g$ & $f_o$ & $f_l$ \\ \hline
        Value & 4 & 0.2 & 0.1 & 30 & 0.065 & 1.5 \\
    \hline
    \end{tabular}
    \caption{Values chosen for the tuning parameters of the proposed algorithm for validation. $w_v$, $w_p$, and $w_I$ are the parameters used in optimal goal selection. $f_g$, $f_o$ and $f_l$ are the parameters used in navigation.}
    \label{tab:goal_select_parameters}
\end{table}
\subsection{Experimental Results}
\begin{figure}[t]
    \centering
    \vspace*{6px}
    \begin{subfigure}{0.45\linewidth}
        \centering
        \includegraphics[width=\linewidth]{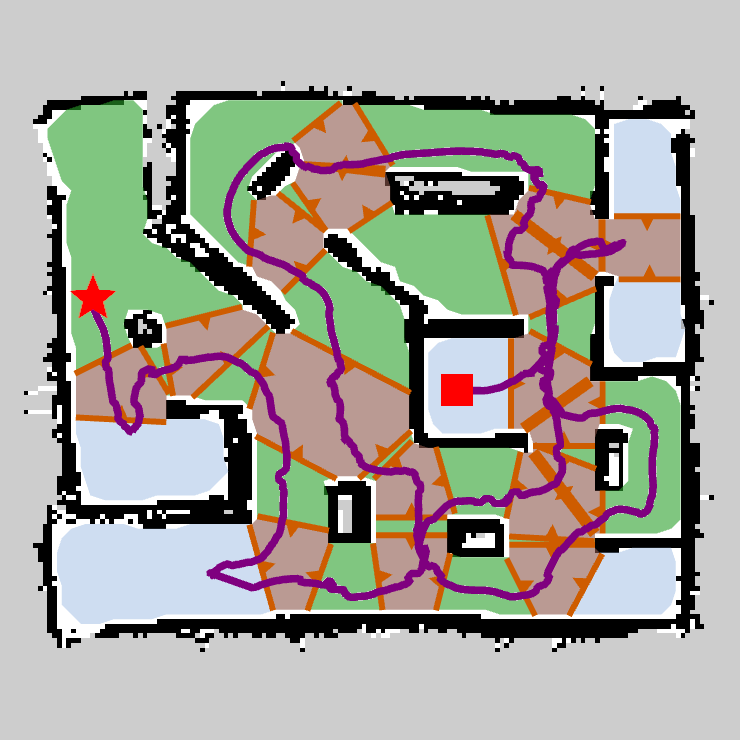}
        \caption{PM Maze 1}
        \label{fig:sub1a}
    \end{subfigure}
    \begin{subfigure}{0.45\linewidth}
        \centering
        \includegraphics[width=\linewidth]{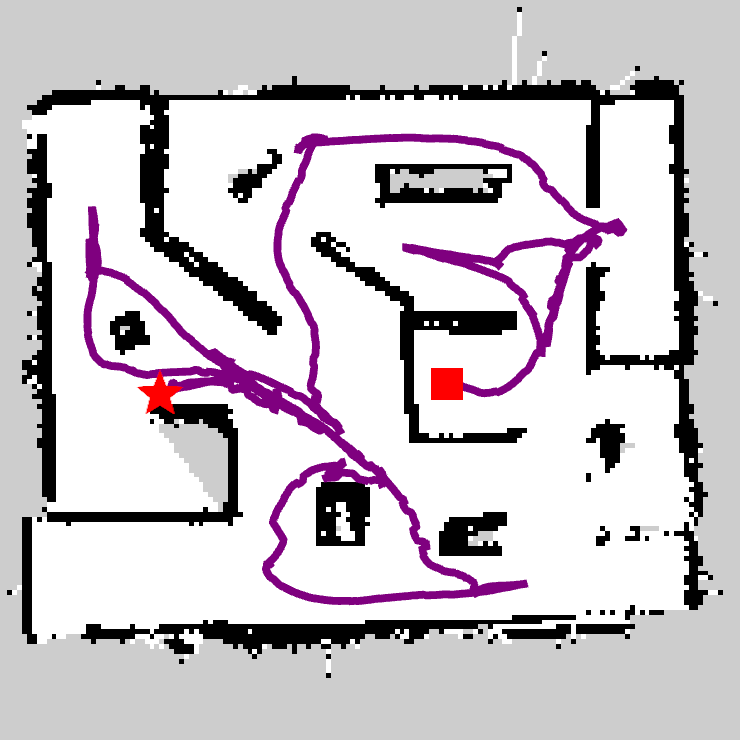}
        \caption{FE Maze 1}
        \label{fig:sub2a}
    \end{subfigure}
    \begin{subfigure}{0.45\linewidth}
        \centering
        \includegraphics[width=\linewidth]{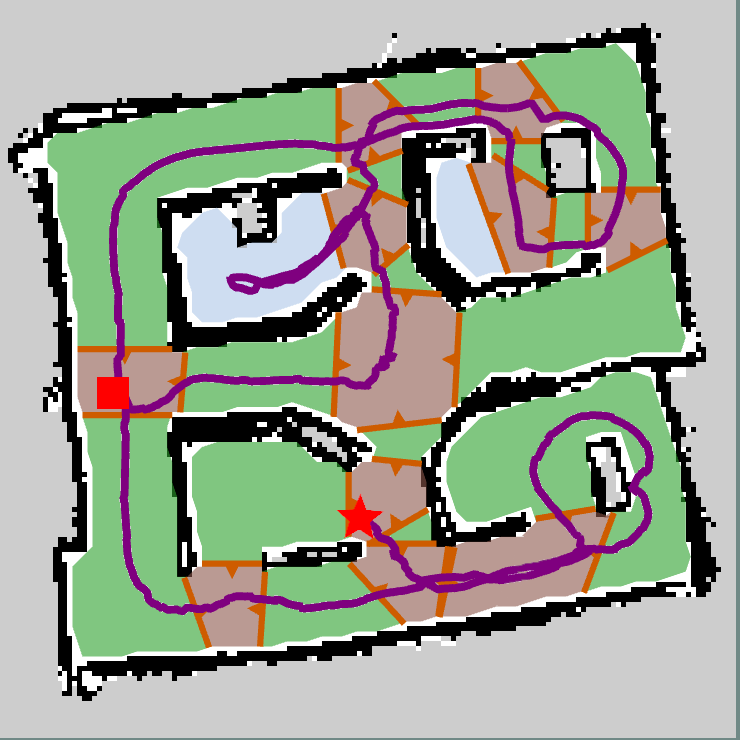}
        \caption{PM Maze 2}
        \label{fig:sub1b}
    \end{subfigure}
    \begin{subfigure}{0.45\linewidth}
        \centering
        \includegraphics[width=\linewidth]{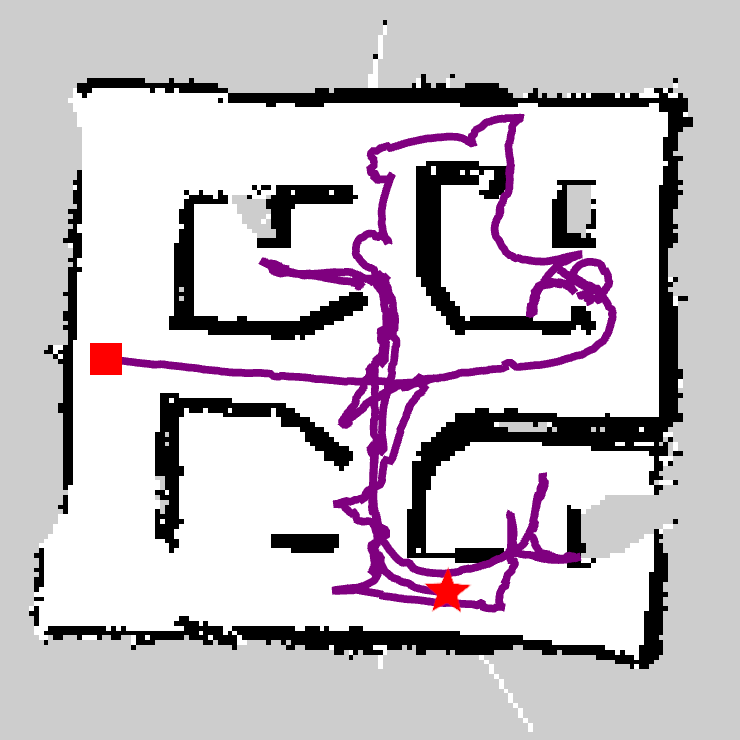}
        \caption{FE Maze 2}
        \label{fig:sub2b}
    \end{subfigure}
    \caption{The maps generated by the run with median exploration time. The starting position of the robot is indicated by a red square, while its final position is denoted by a red star. The path traversed by the robot during its exploration is represented by a purple line. On the map produced by PM the semantic topometric map, generated by the method in \cite{fredriksson2023semantic} is shown as well. 
    }
    \label{fig:runs}
\end{figure}

\begin{figure}[t]
    \centering
    \vspace*{6px}
    \begin{subfigure}{0.49\linewidth}
        \centering
        \includegraphics[width=\linewidth]{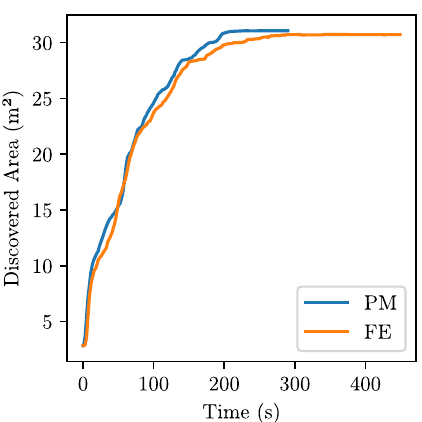}
        \caption{Maze configuration 1}
        \label{fig:timeArea1}
    \end{subfigure}
    \begin{subfigure}{0.49\linewidth}
        \centering
        \includegraphics[width=\linewidth]{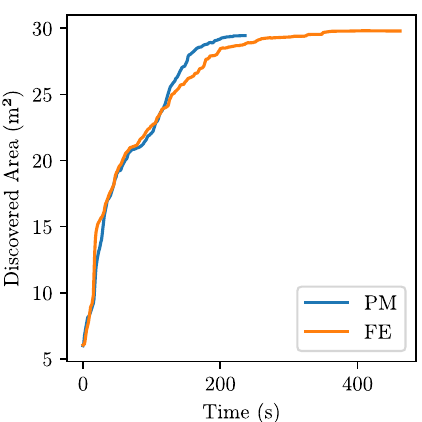}
        \caption{Maze configuration 2}
        \label{fig:timeArea2}
    \end{subfigure}
    \caption{The explored area of unoccupied space (m$^2$) over time, averaged over 10 runs.}
    \label{fig:timeArea}
\end{figure}

\begin{figure}[t]
    \centering
    \includegraphics[width=0.9\linewidth]{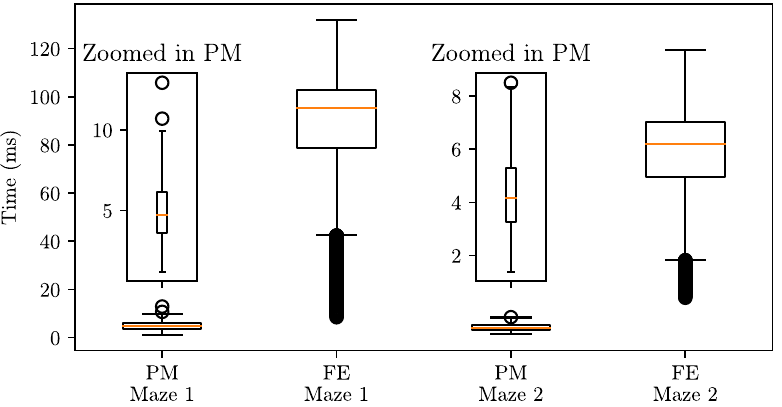}
    \caption{Comparison in compute time (ms) between semantic topological map generation using method in \cite{fredriksson2023semantic} for PM and frontier detection in FE}
    \label{fig:MapGeneration}
\end{figure}

\begin{figure}[t]
    \centering
    \vspace*{12px}
    \includegraphics[width=0.9\linewidth]{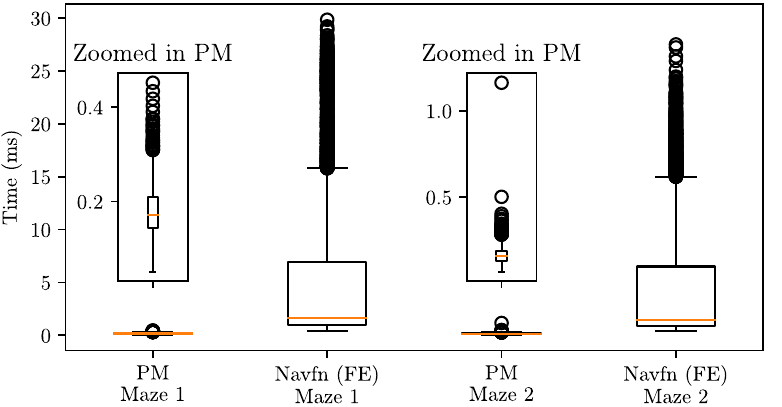}
    \caption{Comparison in compute time (ms) between global navigation for PM and FE}
    \label{fig:GlobalNavigation}
\end{figure}

\begin{figure}[t]
    \centering
    \includegraphics[width=0.9\linewidth]{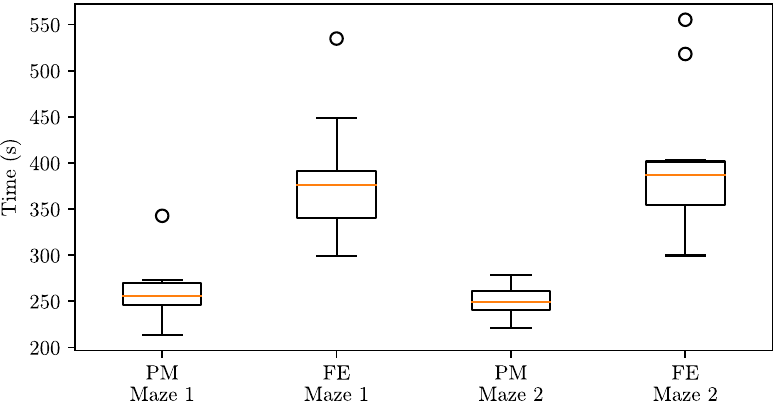}
    \caption{Comparison in exploration mission time (s) between PM and FE.}
    \label{fig:MapExplorationTime}
\end{figure}

The results of the average explored area over time for all ten runs for both PM and FE are depicted in Figure \ref{fig:timeArea} for both test scenarios. The graphs indicate that PM achieves a similar exploration gain to FE at the start of the exploration mission. However, PM explores in a manner that results in less backtracking towards the end of the mission, as can be seen in Figures \ref{fig:runs} and \ref{fig:timeArea}. This results in PM continuing to efficiently explore for longer, resulting in a higher average exploration speed and it finishing its exploration mission faster than FE, as is evident in Figures \ref{fig:timeArea} and \ref{fig:MapExplorationTime}. Another problem that causes the FE method to be slower in exploration time is that it does not check if a frontier is reachable before committing to it, resulting in the robot having to stop its exploration until it can find a frontier it can reach.
Regarding computational performance, Figure \ref{fig:MapGeneration} and \ref{fig:GlobalNavigation} show that both the semantic map generation and navigation used by PM are significantly faster than the frontier detection and navigation adopted by FE. 
\subsection{Comparison and Discussion}
The PM consistently outperformed the FE in terms of exploration speed, time to complete exploration and computational processing time. A significant takeaway from these results is the demonstrated advantage of preprocessing the currently explored map to construct a comprehensive semantic topometric map, rather than merely detecting frontiers, to enable more efficient exploration and navigation. The structural semantics of the explored map hold crucial information for current and future information gain (as demonstrated by the proposed method), which is missed by the state-of-the-art exploration strategies. The noticeable advantages of minimal back-tracking and consistent rise in the information gained throughout the exploration mission are due to the utilization of structural semantic information.

As seen in Figure \ref{fig:GlobalNavigation}, global navigation in the PM is substantially faster than the global planner used by FE. Additionally, the global planner for PM identifies paths to all frontiers, whereas the FE's planner determines only one path at a time. A critical aspect of global planners is their scalability. The FE's planner scales with the map's resolution; in this experiment, the map resolution was 224x224, which is relatively small for an occupancy map. Thus, on a standard or larger sized map, FE's navigation performance is expected to deteriorate generally. In contrast, the PM's global planner scales with the number of intersections, making it more scalable than the FE's planner, as it scales with the complexity (number of intersections) and not the actual size of the map.
An important point of distinction between the semantic-based exploration in PM to existing solutions using semantics is that, while other solutions employ multiple sensors, such as cameras, LiDAR, etc., to detect and classify semantic objects, PM only uses the map generated from SLAM as its input to generate structural semantics. This makes the approach in PM much more fundamental and easier to implement. Another crucial observation is that PM does not utilize any machine learning, which implies that its performance is more reliable in unknown environments compared to machine learning methods that might face issues in environments that are very dissimilar from their training dataset.
\section{CONCLUSIONS}
In this article, we introduced a novel strategy for robotic exploration that utilizes a semantic topometric map. This method departs from traditional frontier-centric approaches by incorporating structural semantic information into the exploration strategy, along with the metric information related to the frontiers. The structural semantics are derived solely from the LiDAR and the occupancy map of the explored parts of the environment by segmenting the map into semantically meaningful areas like intersections, pathways, and dead-ends \cite{fredriksson2023semantic}. 
The proposed method incorporated into optimal goal selection semantic elements such as the distance of the frontier from the nearest intersection and the number of openings out of the intersection. A modified potential field mechanism was also proposed to dynamically adjust path-following based on environmental contexts. This ensures more efficient navigation to compensate for the suboptimal paths in the global skeleton map generated by \cite{fredriksson2023semantic}. Experimental validation demonstrated the superior performance of the proposed method over the \textit{frontier\_exploration} (FE) technique \cite{Topiwala2018} in terms of exploration rate, exploration time, and computational time. In the future, we aim to further investigate the semantic information utilized in the exploration, as well as conduct a more in-depth study on how the different semantic properties influence the exploration process.
\FloatBarrier
\clearpage 
\bibliographystyle{./IEEEtranBST/IEEEtran}
\bibliography{./IEEEtranBST/IEEEabrv,references}

\end{document}